\pdfoutput=1

\documentclass{article}

\PassOptionsToPackage{numbers,compress}{natbib}
\usepackage[final,main]{neurips_2026}

\usepackage{microtype}
\usepackage{graphicx}
\usepackage{subcaption}
\usepackage{booktabs}
\usepackage{amsmath}
\usepackage{amssymb}
\usepackage{mathtools}
\usepackage{amsthm}
\usepackage{xcolor}
\usepackage{bbm}
\usepackage{empheq}
\usepackage{algorithm}
\usepackage{algorithmic}
\usepackage{float}
\usepackage{multirow}
\usepackage{hyperref}
\usepackage[capitalize,noabbrev]{cleveref}

\newcommand{\dd}{\mathrm{d}}

\newcommand{\E}{\mathbb{E}}

\newcommand{\s}{\pi^{(\lambda)}}

\newcommand{\ts}{\theta^{\hat{\pi}^{\vartheta_a}}}

\theoremstyle{plain}
\newtheorem{theorem}{Theorem}[section]

\theoremstyle{definition}

\theoremstyle{remark}

\usepackage[disable,textsize=tiny]{todonotes}

\title{ART for Diffusion Sampling: A Reinforcement Learning Approach to Timestep Scheduling}

\author{
Yilie Huang \\
Department of \\
Applied Mathematics \\
The Hong Kong \\
Polytechnic University \\
Hong Kong \\
\texttt{yilie.huang@polyu.edu.hk}
\And
Wenpin Tang \\
Department of Industrial \\
Engineering and \\
Operations Research \\
Columbia University \\
New York, NY 10027, USA \\
\texttt{wt2319@columbia.edu}
\And
Xun Yu Zhou \\
Department of Industrial \\
Engineering and \\
Operations Research and \\
Data Science Institute \\
Columbia University \\
New York, NY 10027, USA \\
\texttt{xz2574@columbia.edu}
}

\begin{document}
\maketitle

\begin{abstract}
We consider time discretization for score-based diffusion models to generate samples from a learned reverse-time dynamic on a finite grid. Uniform and hand-crafted grids can be suboptimal given a budget on the number of time steps. We introduce Adaptive Reparameterized Time (ART), which controls the clock speed of a reparameterized time variable to redistribute computation along the sampling trajectory while preserving the terminal time, with the objective of minimizing the aggregate Euler discretization error. We derive a randomized companion ART-RL that recasts ART as a continuous-time reinforcement learning problem with Gaussian policies, and prove a two-directional bridge between the two: the deterministic ART optimum lifts to an optimal Gaussian policy, and conversely any optimal Gaussian policy must recover the ART control through its mean. This bridge turns continuous-time actor--critic learning into a principled, rather than heuristic, route to the deterministic timestep optimum. Within the official EDM pipeline, ART-RL improves FID on CIFAR--10 across a wide range of budgets; after one-time offline training, the distilled deterministic schedule transfers without retraining to AFHQv2, FFHQ, and ImageNet at no extra inference cost.
\end{abstract}

\section{Introduction}

Diffusion models \citep{Ho20, Song19, song2020score} are a family of generative models that create samples from unknown target distributions, and underpin recent advances in text-to-image creation \citep{Ramesh22, Rombach2022}, text-to-video generation \citep{Sora2024, Singer2022, Veo2024}, and diffusion large language models \citep{KK25, Nie25}.

A diffusion model consists of two steps:
learn the target score function by training on given samples (pretraining),
and
generate new samples from the learned model (inference).
The focus of this paper is on diffusion sampling/inference,
where a key numerical choice is the timestep grid
used to discretize the learned model.
Standard samplers often use uniform or
hand-crafted timesteps \citep{DDIM, song2020score, karras2022elucidating, Chen2023},
while recent work also studies optimized fixed schedules \citep{sabour2024align}.
Here our purpose is to provide a control-theoretic framework,
which allows strategic and systematic time discretization in diffusion sampling in order to minimize the aggregate error when applying the Euler scheme.
The main contribution of this paper is as follows:
\begin{itemize}
\item
{\em  Formulation and methodology}:
We propose an optimal control framework -- {\em Adaptive Reparameterized Time} (ART),
to schedule timesteps for diffusion sampling.
The idea is to treat the speed of the diffusion sampler as control
to reparameterize time and adaptively redistribute computation along the sampling trajectory.
To solve the ART problem,
we propose a continuous-time reinforcement learning (CTRL) approach -- ART-RL,
premised upon the recent development in \citet{wang2020reinforcement, jia2021policy, jia2021policypg}.
\item
{\em  Theory}:
We establish a two-directional bridge between ART and ART-RL: \emph{sufficiency} shows that the deterministic ART optimum lifts to an optimal Gaussian policy whose mean is exactly the ART control, and \emph{necessity} shows that any optimal Gaussian policy must recover the ART control through its mean. This justifies parameterizing the actor as a Gaussian mean and deploying it deterministically with rigorous guarantees, and yields actor--critic updates for learning adaptive time schedules.
\item
{\em Experiments}:
Our ART-RL schedule consistently outperforms Uniform, the DPM-Solver log-SNR grid \citep{lu2022dpm}, and EDM schedules \citep{karras2022elucidating} in a matched CIFAR--10 EDM evaluation. It improves FID across a broad range of sampling time budgets, with large gains at low and moderate budgets and a consistent advantage at the largest budget.
\item
{\em Generalization and amortization}:
ART-RL separates offline schedule learning from inference-time sampling.
After training, the stochastic policy is distilled into a deterministic time-only grid that can be reused as a drop-in schedule across timestep counts and transferred across various datasets without retraining.
\end{itemize}

To the best of our knowledge, this is the first control/RL formulation of diffusion timestep scheduling through adaptive time reparameterization, and the first to provide a two-directional bridge between a deterministic timestep optimum and an actor--critic learning procedure. The proposed method is data-driven and exhibits robust performance with reusable, amortized transfer.

\medskip
\noindent
{\bf Relevant literature}:
Diffusion models as generative tools
were first proposed by \citet{Ho20} (DDPM) and \citet{DDIM} (DDIM) in the discrete setting.
The pioneering work of \citet{song2020score} introduces a continuous-time formulation of diffusion models,
providing a unified treatment that encompasses earlier discrete-time models.
Various sampling methods have been proposed for diffusion inference,
including predictor-corrector sampler \citep{song2020score}, exponential integrator \citep{ZC23},
higher-order solvers \citep{ZSC23, WCW24}, and optimized fixed timestep schedules \citep{sabour2024align}.
See also \citet{LLT22, Chen2023, ChenChe23, LW23, Ben24, LYl24, HHL25}
for convergence analysis of diffusion models (with either uniform or hand-crafted timesteps).

Continuous-time reinforcement learning (CTRL) was first formulated by \citet{wang2020reinforcement}, where exploration is modeled via stochastic relaxed control to capture the trial-and-error nature of reinforcement learning. Subsequent work developed a model-free theoretical foundation via martingale methods \citep{jia2021policy, jia2021policypg, JZ22, TZq24}, established performance guarantees \citep{huang2025alm,huang2025sublinear,huang2024mean}, and studied policy optimization \citep{ZTY23}. The CTRL framework has also been applied to training and fine-tuning diffusion models in generative AI \citep{GZZ24, huang2026aid, ZC24, ZC25}.

\medskip
\noindent
{\bf Organization}:
Section~\ref{sc2} reviews diffusion models and introduces the ART control formulation. Section~\ref{sc3} describes how CTRL is used to solve ART, leading to the ART-RL approach. Section~\ref{sc4} presents the ART-RL algorithm, followed by empirical results in Section~\ref{sec:experiments}. Section~\ref{sc6} concludes. An extended ART--ART-RL analysis, including all proofs, and additional numerical results appear in the extended version \citep{huang2026full}.

\section{Diffusion Models and ART}
\label{sc2}

\subsection{Continuous-Time Score-Based Diffusion Models}

We briefly review continuous-time score-based diffusion models; see \citet{TZSS25} for a recent survey. Given samples from an unknown target distribution, a forward process corrupts data over time \(\tau\in[0,T]\) toward a tractable reference distribution, and sampling proceeds by integrating a learned reverse-time process.


\paragraph{Diffusion sampling.}
A forward process obeys the It\^o SDE
\begin{equation}\label{eq:forward-sde}
\mathrm{d}\bar{x}(\tau)=-f(\tau)\bar{x}(\tau)\,\mathrm{d}\tau+g(\tau)\,\mathrm{d}w(\tau),
\end{equation}
where \(\tau\in[0,T]\), \(\bar{x}(0)\sim p_0\) is the target data distribution, $w$ is a standard Wiener process in $\mathbb{R}^d$, and $f$ and $g$ are coefficient functions. Let $p_\tau$ be the law of $\bar{x}(\tau)$ and  $S(\tau,x)=\nabla_x\log p_\tau(x)$ be the score.

For sampling, we use the probability flow ODE associated with \eqref{eq:forward-sde},
which shares the same marginals as the reverse-time SDE; see e.g. \citet[Theorem 5.1]{TZSS25}. Denote the backward state by $\tilde{x}(\tau)\coloneqq \bar{x}(T-\tau)$ with initialization $\tilde{x}(0)\sim p_T$. Replacing $S$ with a trained score model $\hat{S}$ yields
\begin{equation}\label{eq:backward-pf}
\frac{\mathrm{d}\tilde{x}(\tau)}{\mathrm{d}\tau}=f(T-\tau)\tilde{x}(\tau)+\tfrac12 g(T-\tau)^2\hat{S}(T-\tau,\tilde{x}(\tau)),
\end{equation}
where \(\tau\in[0,T]\).

\paragraph{Euler discretization.}
We integrate \eqref{eq:backward-pf} on a grid $0=\tau_0<\tau_1<\cdots<\tau_K=T$ with step sizes $h_i=\tau_{i+1}-\tau_i$, and write $\tilde{x}_i\coloneqq \tilde{x}(\tau_i)$. The explicit Euler update is
\begin{equation}
\tilde{x}_{i+1}=\tilde{x}_i+h_i\left[f(T-\tau_i)\tilde{x}_i+\tfrac12 g(T-\tau_i)^2\hat{S}(T-\tau_i,\tilde{x}_i)\right].
\label{eq:tau-discretization}
\end{equation}

A uniform grid $\tau_i = iT/K$ is simple but allocates evaluations uniformly even when the reverse dynamics vary along the trajectory. A simple intuition is that early stages, where samples resemble noise, may tolerate coarser steps, while later stages will benefit from finer resolutions. Such considerations motivate adaptive, data-driven schedules that redistribute steps under a given, fixed total time budget $T$.

\subsection{ART: Time Reparameterization via Control}

Motivated by the drawback of uniform timesteps in backward sampling discussed above, we introduce a reparameterized sampling clock and formalize adaptive timestep selection in the reverse process as a control problem. The key idea is to replace the uniform progression of original time by an adaptive, controlled progression that can accelerate in some segments and decelerate in others, thereby redistributing computational effort along the trajectory.

To this end, let $\psi:[0,T]\to\mathbb{R}$ be a continuous time mapping from the new clock $t$ to the original diffusion time $\tau$ (i.e., $\tau=\psi(t)$), with $\psi(0)=0$ and $\psi(T)=T$.
On this new clock we write the state as $x(t)\coloneqq \tilde{x}(\psi(t))$, namely the backward state evaluated at diffusion time $\psi(t)$, with $x(0)\sim p_T$.
We define the control as $\theta(t)\coloneqq \dot{\psi}(t)$, which quantifies the instantaneous rate of change in diffusion time with respect to $t$ and satisfies
$\int_{0}^{T}\theta(t)\,\mathrm{d}t=\psi(T)-\psi(0)=T$.
To keep the formulation general, we do not impose monotonicity of $\psi$, so $\theta$ may take either sign.
The monotone time-change case (equivalently, $\theta(t)\ge 0$ almost everywhere) is included as a special case of this formulation. 
In particular, if we discretize the new clock uniformly as $0=t_0<t_1<\cdots<t_K=T$, then the induced original-time grid is $\tau_i\coloneqq \psi(t_i)$ with nonuniform step sizes $\tau_{i+1}-\tau_i$. Intuitively, $x(\cdot)$ tracks the generative trajectory under the reparameterized time, while $\theta(\cdot)$ determines how quickly the trajectory advances in original time per unit step on the new clock (larger $\theta(t)$ corresponds to faster progression at $t$). We refer to this time reparameterized sampling framework as \emph{Adaptive Reparameterized Time} (ART).

By the chain rule, ART induces the controlled dynamics
\begin{subequations}\label{eq_original_dynamics}
\begin{empheq}[left=\empheqlbrace]{align}
\dot{x}(t)&=\theta(t)F(x(t),\psi(t)),\quad x(0)\sim p_T,
\label{eq:reparam-dyn-x}\\
\dot{\psi}(t)&=\theta(t),\quad \psi(0)=0,\ \psi(T)=T,
\label{eq:reparam-dyn-psi}
\end{empheq}
\end{subequations}
where the backward probability–flow vector evaluated at original time $T-\psi$ is
\begin{equation}\label{eq:F-def}
F(x,\psi)\coloneqq f(T-\psi)x+\tfrac12 g(T-\psi)^2\hat{S}(T-\psi,x).
\end{equation}
The constraint \(\int_{0}^{T}\theta(t)\,\mathrm{d}t=T\) is the time budget allocated by the control.

To assess how Euler behaves on the $t$-clock, we relate its local approximation error over a step to the curvature of the right-hand side in \eqref{eq:reparam-dyn-x}. We proceed analogously to the Euler discretization in \eqref{eq:tau-discretization}. On an interval $t\in[t_i,t_{i+1})$ with stepsize $h_i:=t_{i+1}-t_i$, the Euler update uses one control value per step; we denote this step value by $\theta_i$.

Define the one-step Euler error proxy on the $t$-clock by \(E_i:=x(t_{i+1})-\{x(t_i)+h_i\theta_iF(x(t_i),\psi(t_i))\}\). A Taylor expansion yields \(E_i=\frac{h_i^2}{2}\theta_i^2Q(x(t_i),\psi(t_i))+O(h_i^3)\),
where
\begin{equation}\label{eq:Qexplicit}
Q(x,\psi)=A(s,x)B(s,x)-g(s)g'(s)\hat S(s,x)-f'(s)x-\tfrac12 g(s)^2\,\partial_s\hat S(s,x),
\end{equation}
with
\( s:=T-\psi \),
\( A(s,x):=f(s)I_d+\tfrac12 g(s)^2\nabla_x\hat S(s,x) \),
and
\( B(s,x):=f(s)x+\tfrac12 g(s)^2\hat S(s,x) \).
Thus, to leading order, the magnitude of the Euler local error on step $i$ is proportional to
$\theta_i^2|Q(x(t_i),\psi(t_i))|$, where $|\cdot|$ denotes the Euclidean norm throughout. In implementation, $\nabla_x\hat S(s,x)$ is not formed; automatic differentiation provides the needed Jacobian--vector product.

This motivates using $|Q(x,\psi)|\,\theta(t)^2$ as an error-density surrogate for allocating the time-warping rate. With a fixed time budget, we choose $\theta=\theta(\cdot)$ to minimize the overall residual surrogate subject to the constraint $\int_0^T \theta(t)\,\mathrm{d}t = T$, enforced via a Lagrange multiplier $\gamma \in \mathbb{R}$. Equivalently, we maximize the negative Lagrangian cost
\begin{equation}\label{eq_original_objective}
J^\theta(s,y,\phi)=\E\Big[\int_s^T\!\bigl(-|Q(x(t),\psi(t))|\theta^2(t)-\gamma\theta(t)\bigr)\,\dd t+\gamma T \,\Bigm|\, x(s)=y,\ \psi(s)=\phi \Big].
\end{equation}

Denote the optimal value function to be
\begin{equation}\label{eq_original_value_function}
V(s,y,\phi)=\max_{\theta=\theta(\cdot)}J^\theta(s,y,\phi).
\end{equation}

In summary, ART recasts timestep allocation as continuous-time control of the time-warping rate \(\theta\) under \eqref{eq_original_dynamics} and \eqref{eq_original_objective}. The next section develops an RL-based procedure (ART–RL) to learn \(\theta\) in a data-driven way.

\section{Randomized Control and RL Formulation}
\label{sc3}

The ART problem has no closed-form solution and its HJB is intractable on the high-dimensional state space \(x\in\mathbb{R}^{d}\). We therefore introduce a Gaussian randomization of the control as a technical device that embeds ART into the continuous-time RL framework, enabling actor--critic learning. We call this approach \emph{Adaptive Reparameterized Time via Reinforcement Learning} (ART-RL).

\subsection{ART-RL: Auxiliary Problem with Gaussian Policies}
\label{sec:art-rl}
We model control randomization with a stochastic policy over the time-warping rate $\theta$. Inspired by entropy-regularized control \citep{ziebart2008maximum,huang2022achieving}, we consider the Gaussian family indexed by \(\lambda\ge0\):
\begin{equation}
\label{eq:gaussian_policy}
\s(\cdot \mid t,x,\psi)=\mathcal{N}\!\left(\mu(t,x,\psi),\,\frac{\lambda}{\lvert Q(x,\psi)\rvert}\right),
\end{equation}
where $\mu$ is a measurable deterministic policy and $Q$ is defined in~\eqref{eq:Qexplicit}. Let $\Pi^{(\lambda)}$ be the class of such policies with finite second moments. Since \(Q\) is proportional to the Euler local-error proxy, the variance is smaller in stiff regions and larger elsewhere; \(\lambda\) controls the overall noise level without changing the mean. For analysis we assume \(\lvert Q(x,\psi)\rvert>0\) almost surely on compact intervals; in practice we use \(\lvert Q\rvert\vee\varepsilon\) for small \(\varepsilon>0\).

Given a policy $\s\in\Pi^{(\lambda)}$, let $\bigl(x^{\s}(t),\psi^{\s}(t)\bigr)_{t\in[0,T]}$ denote the corresponding state process which, according to \citet{wang2020reinforcement}, satisfy
\begin{subequations}\label{eq_auxiliary_dynamics}
\begin{empheq}[left=\empheqlbrace]{align}
\dot x^{\pi^{(\lambda)}}(t)&=\!\int_{\mathbb{R}}\!\theta\,F(x^{\pi^{(\lambda)}}(t),\psi^{\pi^{(\lambda)}}(t))\,\s_t(\theta)\,\dd\theta,\quad x^{\pi^{(\lambda)}}(0)=x_0\sim p_T,
\label{eq:aux_dynamics_x}\\
\dot\psi^{\pi^{(\lambda)}}(t)&=\!\int_{\mathbb{R}}\!\theta\,\s_t(\theta)\,\dd\theta,\quad \psi^{\pi^{(\lambda)}}(0)=0,\ \psi^{\pi^{(\lambda)}}(T)=T.
\label{eq:aux_dynamics_psi}
\end{empheq}
\end{subequations}

The corresponding total-time constraint is $\int_0^T\int_{\mathbb{R}}\theta\s_t(\theta)\dd\theta\dd t=T$. The objective function for the auxiliary problem is
\[
\begin{aligned}
J^{\s}(s,y,\phi)&=\E\Big[\gamma T+\lambda T+\int_s^T\int_{\mathbb{R}}\s_t(\theta)\bigl(-|Q(x^{\s}(t),\psi^{\s}(t))|\theta^2-\gamma\theta\bigr)\dd\theta\,\dd t \\
&\qquad\qquad\qquad\qquad\qquad\qquad\Bigm|\,x^{\s}(s)=y,\ \psi^{\s}(s)=\phi\Big],
\end{aligned}
\]
where the Lagrange multiplier $\gamma\in\mathbb{R}$ enforces the total-time constraint and $\lambda\ge 0$ is the variance parameter from \eqref{eq:gaussian_policy}. The optimal value function is

\begin{equation}\label{eq_auxiliary_value_function}
V^{(\lambda)}(s,y,\phi) \;=\; \max_{\s\in\Pi^{(\lambda)}} J^{\s}(s,y,\phi).
\end{equation}

\subsection{Relationship between ART and ART-RL}

We now establish the relationship between the ART control problem \eqref{eq_original_value_function} and the ART-RL auxiliary problem \eqref{eq_auxiliary_value_function}. By dynamic programming, the optimal value function \(V\) for \eqref{eq_original_value_function} satisfies
\begin{equation}
\label{eq_original_hjb}
V_t + \sup_{\theta} \biggl\{ (V_x^\top F(x, \psi) + V_{\psi} - \gamma) \theta - |Q(x, \psi)| \theta^2 \biggr\} = 0,
\end{equation}
with terminal condition \(V(T,x,\psi)=\gamma T\).

Moreover, by \citet{wang2020reinforcement}, the optimal value function \(V^{(\lambda)}\) of \eqref{eq_auxiliary_value_function} satisfies the exploratory HJB:
\begin{equation}\label{eq_auxiliary_hjb}
V^{(\lambda)}_t+\sup_{\mu}\Bigl\{\bigl(V^{(\lambda)\top}_x F(x,\psi)+V^{(\lambda)}_{\psi}-\gamma\bigr)\mu-|Q(x,\psi)|\!\left(\mu^2+\frac{\lambda}{|Q(x,\psi)|}\right)\Bigr\}=0
\end{equation}
with terminal condition \(V^{(\lambda)}(T,x,\psi)=(\gamma + \lambda) T\).

The following three results characterize the connection: a value-shift theorem aligning the deterministic and auxiliary HJB equations, followed by sufficiency and necessity theorems that together establish a two-directional bridge.
All proofs can be found in the extended version \citep{huang2026full}.
\begin{theorem}[Value function shift]
\label{thm_value_shift}
If \(V\) is a classical solution of \eqref{eq_original_hjb}, then
\[
V^{(\lambda)}(t,x,\psi) = V(t,x,\psi) + \lambda t
\]
is a classical solution of \eqref{eq_auxiliary_hjb}.
\end{theorem}

\begin{theorem}[Sufficiency: deterministic ART optimum lifts to an optimal Gaussian policy]
\label{thm_policy_recovery}
Under the conditions of Theorem~\ref{thm_value_shift}, define
\[
\mu^*(t,x,\psi)=\frac{V_x^\top F(x,\psi)+V_{\psi}-\gamma}{2|Q(x,\psi)|}.
\]
Then the Gaussian policy
\[
\pi^{(\lambda)*}(\cdot \mid t,x,\psi)=\mathcal{N}\!\left(\mu^*(t,x,\psi),\,\frac{\lambda}{|Q(x,\psi)|}\right)
\]
is optimal for the auxiliary problem \eqref{eq_auxiliary_value_function} subject to the dynamics \eqref{eq_auxiliary_dynamics}. Furthermore, \(\mu^*\) is an optimal policy for the original problem \eqref{eq_original_value_function} subject to the dynamics \eqref{eq_original_dynamics}.
\end{theorem}

\begin{theorem}[Necessity: optimal auxiliary policy recovers the ART control]
\label{thm_necessity_mean}
Suppose Theorem~\ref{thm_value_shift} holds, \(|Q(x,\psi)|>0\), and the policy-gap residual \(R^{\pi^{(\lambda)}}(t,x,\psi)=|Q(x,\psi)|(\mu^{\pi^{(\lambda)}}(t,x,\psi)-\mu^*(t,x,\psi))^2\) is continuous in \((t,x,\psi)\). Let \(\pi^{(\lambda),\ast}\in\Pi^{(\lambda)}\) be any Gaussian policy that is optimal for the auxiliary problem \eqref{eq_auxiliary_value_function} at every initial pair \((s,y,\phi)\). Then its mean field satisfies
\[
\mu^{\pi^{(\lambda),\ast}}(t,x,\psi)\;=\;\mu^{*}(t,x,\psi)\qquad \text{for every }(t,x,\psi),
\]
where \(\mu^{*}\) is the optimal ART control of Theorem~\ref{thm_policy_recovery}. In particular, \(\mu^{\pi^{(\lambda),\ast}}\) is also an optimal control for the original ART problem~\eqref{eq_original_value_function}.
\end{theorem}

Theorems~\ref{thm_value_shift}--\ref{thm_necessity_mean} establish a \emph{two-directional} equivalence between ART and ART-RL. Hence the actor in Section~\ref{sc4} can be parameterized as a Gaussian mean and deployed deterministically, with the guarantee that any optimal auxiliary policy recovers the ART control through its mean. This bridge is what makes continuous-time actor--critic learning a principled, rather than heuristic, route to the deterministic timestep problem.

\section{ART-RL Actor--Critic Algorithm}\label{sec_algorithm}
\label{sc4}

Theorems~\ref{thm_value_shift}--\ref{thm_necessity_mean} reduce ART to solving the auxiliary ART-RL problem and guarantee that any optimal auxiliary policy mean recovers the deterministic ART optimum.

\subsection{Algorithm Design}\label{sec_alg}
Our algorithm builds on the continuous-time actor--critic framework of \citet{jia2021policypg}, adapted to ART-RL with randomized Gaussian control. We use two neural networks \(NN^{\vartheta_c}\) and \(NN^{\vartheta_a}\) for the value function and policy mean, and parameterize
\begin{equation}\label{eq_parameterization}
\hat{V}^{\vartheta_c}(t,x,\psi)=NN^{\vartheta_c}(t,x,\psi)+\lambda t,\qquad
\hat{\pi}^{\vartheta_a}(\cdot \mid t,x,\psi)=\mathcal{N}\!\left(NN^{\vartheta_a}(t,x,\psi), \frac{\lambda}{|Q(x,\psi)|}\right).
\end{equation}
Let \(\ts\sim\hat{\pi}^{\vartheta_a}\) be a Gaussian control process with corresponding (observable) state process \((x^{\ts}(t),\psi^{\ts}(t))\). For brevity, in \eqref{eq_moment_conditions} we write \((x(t),\psi(t),\theta(t))=(x^{\ts}(t),\psi^{\ts}(t),\ts(t))\). By \citet{jia2021policypg}, \(\hat{V}^{\vartheta_c}\) and \(\hat{\pi}^{\vartheta_a}\) satisfy the coupled moment conditions
\begin{equation}\label{eq_moment_conditions}
\begin{aligned}
\mathbb{E}\biggl[\int_0^T \frac{\partial NN^{\vartheta_c}(t,x(t),\psi(t))}{\partial\vartheta_c}\Bigl(\dd\hat{V}^{\vartheta_c}(t,x(t),\psi(t))-\bigl(|Q(x(t),\psi(t))|\theta(t)^2+\gamma\theta(t)\bigr)\dd t\Bigr)\biggr]&=0,\\
\mathbb{E}\biggl[\int_0^T \frac{\partial\log\hat{\pi}^{\vartheta_a}(\theta(t)\mid t,x,\psi)}{\partial\vartheta_a}\Bigl(\dd\hat{V}^{\vartheta_c}(t,x(t),\psi(t))-\bigl(|Q(x(t),\psi(t))|\theta(t)^2+\gamma\theta(t)\bigr)\dd t\Bigr)\biggr]&=0.
\end{aligned}
\end{equation}
These are standard moment conditions in RL \citep{sutton1998reinforcement,huang2025data}; the parameters $\vartheta_c$ and $\vartheta_a$ are updated by stochastic approximation.

\subsection{Implementable ART-RL Algorithm}\label{sec:pseudocode}
At iteration \(n\), a trajectory \((x_{n},\psi_{n},\theta_{n})\) is generated under the current policy \(\hat{\pi}^{\vartheta_{a,n}}\) on a uniform grid \(0=t_{0}<\cdots<t_{K}=T\) with \(\Delta t=T/K\), and parameters are updated with learning rate \(a_{n}>0\). Writing \(\hat{V}^{\vartheta_{c,n}}_{k}:=\hat{V}^{\vartheta_{c,n}}(t_{k},x_{n}(t_{k}),\psi_{n}(t_{k}))\) and applying stochastic approximation with Riemann discretization to the moment conditions \eqref{eq_moment_conditions} gives, with \(D_{n,k}:=\hat{V}^{\vartheta_{c,n}}_{k+1}-\hat{V}^{\vartheta_{c,n}}_{k}-\gamma_{n}\theta_{n}(t_{k})\Delta t-|Q(x_{n}(t_{k}),\psi_{n}(t_{k}))|\theta_{n}(t_{k})^{2}\Delta t\), the critic and actor updates
\begin{subequations}\label{eq_disc_updating_rules}
\begin{align}
\label{eq_disc_critic_update}
\vartheta_{c,n+1}&\leftarrow\vartheta_{c,n}+a_n\sum_{k=0}^{K-1}\frac{\partial NN^{\vartheta_{c,n}}}{\partial\vartheta_c}\bigl(t_k,x_n(t_k),\psi_n(t_k)\bigr)D_{n,k}, \\
\label{eq_disc_actor_update}
\vartheta_{a,n+1}&\leftarrow\vartheta_{a,n}+a_n\sum_{k=0}^{K-1}\frac{\partial\log \hat{\pi}^{\vartheta_{a,n}}_{t_k}(\theta_n(t_k))}{\partial\vartheta_a}D_{n,k},
\end{align}
\end{subequations}
where \(\hat{\pi}^{\vartheta_{a,n}}_{t_{k}}(\theta):=\hat{\pi}^{\vartheta_{a,n}}(\theta\mid t_{k},x_{n}(t_{k}),\psi_{n}(t_{k}))\). The Lagrange multiplier is updated by \(\gamma_{n+1}\leftarrow\gamma_{n}+a_{n}\bigl(\psi_{n}(T)-T\bigr)\), where \(\psi_{n}(T)=\psi_{n}(t_{K})\). The full procedure is summarized in Algorithm~\ref{alg:art-rl-actor-critic}.
\begin{algorithm}[htb]
\caption{Time-discretized ART-RL Actor-Critic}\label{alg:art-rl-actor-critic}
\begin{algorithmic}
\FOR{$n = 1$ to $N$}
    \STATE Set $k = 0$, $t = t_k = 0$, initialize $(x_n(t_0), \psi_n(t_0))$
    \WHILE{$t < T$}
        \STATE Sample control \(\theta_n(t_k)\) from the Gaussian policy \eqref{eq_parameterization}
        \STATE Update the states  by the ART dynamics \eqref{eq_original_dynamics}
        \STATE Increment time: $t_{k+1} = t_k + \Delta t$, $k \leftarrow k+1$
    \ENDWHILE
    \STATE Collect trajectory $\{(t_k, x_n(t_k), \psi_n(t_k), \theta_n(t_k))\}_{k=0}^{K-1}$
    \STATE Update critic and actor parameters via \eqref{eq_disc_critic_update} and \eqref{eq_disc_actor_update}
    \STATE Update the Lagrange multiplier \(\gamma_{n+1}\)
\ENDFOR
\end{algorithmic}
\end{algorithm}

\section{Numerical Experiments}\label{sec:experiments}
We evaluate ART-RL in a one-dimensional known-score setting and in image-generation experiments using the official EDM pipeline \citep{karras2022elucidating}. Across CIFAR--10, AFHQv2, FFHQ, and ImageNet, the score model, solver, and EDM implementation are kept fixed and only the timesteps are replaced. We compare Uniform, the DPM-Solver log-SNR grid, EDM timesteps, and ART-RL, reporting \(W_{2}\) on the 1D problem and FID vs.\ NFE on images; on CIFAR--10 we additionally include AYS \citep{sabour2024align}. Additional experimental details and results are in the extended version \citep{huang2026full}.

\begin{figure}[t]
  \centering
  \includegraphics[width=\linewidth]{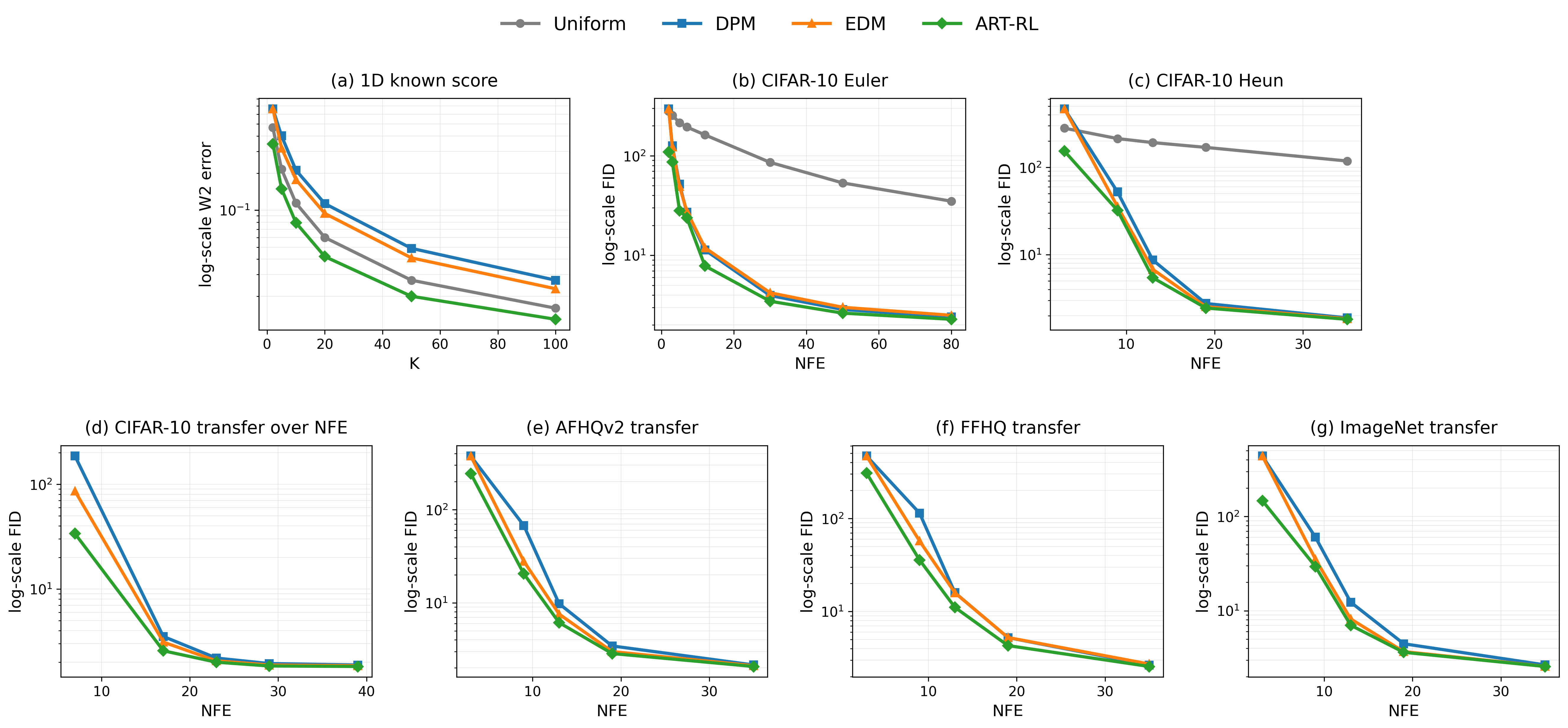}
  \caption{Quantitative overview: ART-RL Pareto-dominates Uniform, DPM-logSNR, and EDM on CIFAR--10 across NFE budgets, and the same distilled grid retains the advantage on AFHQv2, FFHQ, and ImageNet without retraining. DPM-logSNR is the DPM-Solver uniform log-SNR grid.}
  \label{fig:quant-overview}
\end{figure}

\subsection{One--Dimensional Example with Analytical Score}\label{subsec:oned}
We first consider a one-dimensional example where the score function is analytical in closed form; hence every performance difference comes from the timestep schedule rather than score-estimation error. The forward process starts from \(p_0=\mathcal{N}(0,1)\) and follows \(\mathrm{d}x(t)=\sqrt{2t}\,\mathrm{d}w(t)\) over \(T=3\), so \(x(t)\sim\mathcal{N}(0,1+t^2)\), \(p_T=\mathcal{N}(0,10)\), and \(S(t,x)=-x/(1+t^2)\). Substituting this score into \eqref{eq:F-def} and \eqref{eq:Qexplicit} gives \(F(x,\psi)=-(T-\psi)x/(1+(T-\psi)^2)\) and \(Q(x,\psi)=x/(1+(T-\psi)^2)^2\).

This controlled setting separates learned from hand-designed schedules: EDM and DPM-logSNR are strong but fixed analytic prescriptions, whereas ART-RL learns the time allocation from the objective. In Figure~\ref{fig:quant-overview}-(a), ART-RL lowers \(W_2\) across all tested \(K\); for example, at \(K=10\), ART-RL obtains $.079$ versus $.177$ (EDM) and $.211$ (DPM-logSNR), and at \(K=100\), $.013$ versus $.023$ and $.027$.

\paragraph{Distillation diagnostic.}
Training trajectories concentrate tightly around a smooth mean control, so the stochastic policy can be distilled into a deterministic time-only grid, with two practical benefits. First, distillation removes the per-step cost of actor and \(Q\) evaluations, so ART-RL sampling requires no extra computation beyond standard schemes such as Uniform or EDM. Second, it eliminates the residual \(\psi(T)\) overshoot/undershoot that occurs when \(\theta\) is produced by a neural-network actor: normalizing the deterministic increments to sum exactly to \(T\) guarantees the induced grid hits the terminal time precisely, improving the numerical fidelity of the discretized probability flow ODE.

\subsection{CIFAR--10 under the EDM Pipeline}\label{subsec:cifar}
We first run a solver-aligned Euler ablation to verify the objective--sampler alignment. ART-RL improves FID over Uniform, DPM-logSNR, and EDM across all Euler budgets; for example, at \(\mathrm{NFE}=30\) it attains \(3.46\) versus \(3.95\) (DPM-logSNR), \(4.21\) (EDM), and \(85.83\) (Uniform).

We report the main image results with the default Heun solver. Heun is the sampler used in the official EDM pipeline and a 2nd-order Runge--Kutta method (one order higher than Euler), which makes it both the standard choice for modern low-NFE deterministic diffusion sampling and the configuration in which EDM reports its strongest FID; running under Heun therefore tests ART-RL against the strongest baseline setup. Although our training surrogate is Euler-based, we assess transfer of the Euler-trained schedule under Heun empirically below. We test \(K\in\{2,5,7,10,18\}\) with Euler only for the final step update, matching the EDM implementation and giving \(\mathrm{NFE}=2K-1\). Table~\ref{tab:cifar10} shows that ART-RL attains the lowest FID across all baselines. The improvement is most pronounced at small and moderate budgets, and remains present at the largest-budget setting where EDM reported its strongest performance: at \(\mathrm{NFE}=35\), ART-RL obtains \(1.82\), compared with \(1.85\) for EDM and \(1.89\) for DPM-logSNR.

\begin{table}[htbp]
\centering
\caption{CIFAR--10 FID vs.\ NFE. DPM-logSNR and EDM use fixed analytic schedules, whereas AYS is an optimized fixed reference schedule. At \(\mathrm{NFE}=3\), all three use the same boundary noise levels and therefore yield the same FID; ART-RL instead learns its schedule from data.}
\label{tab:cifar10}
\begin{tabular}{lrrrrr}
\toprule
NFE & 3 & 9 & 13 & 19 & 35 \\
\midrule
Uniform & 280.29 & 213.13 & 191.69 & 168.87 & 118.02 \\
AYS & 465.83 & 157.29 & 86.82 & 35.30 & 3.60 \\
DPM-logSNR & 465.83 & 52.29 & 8.67 & 2.76 & 1.89 \\
EDM & 465.83 & 35.54 & 6.79 & 2.54 & 1.85 \\
ART-RL & \textbf{152.86} & \textbf{32.13} & \textbf{5.44} & \textbf{2.45} & \textbf{1.82} \\
\bottomrule
\end{tabular}
\end{table}

Visual comparisons corroborate the table: at very small budgets ART-RL already produces recognizable objects while EDM and DPM-logSNR remain close to noise.

\paragraph{Statistical significance.}
At \(\mathrm{NFE}=35\), three matched 50{,}000-sample runs give ART-RL FIDs \(1.82,1.79,1.82\) versus EDM FIDs \(1.85,1.83,1.85\); ART-RL's worst run beats EDM's best. For context, the ART-RL vs.\ EDM gaps on CIFAR--10 are \(313\), \(3.4\), and \(1.35\) at \(\mathrm{NFE}=3,9,13\), respectively; these selected gaps are larger than the seed-to-seed 50k-sample FID standard deviation reported by EDM (\(\approx 0.02\)--\(0.05\)).

\subsection{Generalization of the ART-RL Time Schedule}\label{sec:generalization}
The ART-RL schedule is learned once on CIFAR--10 (1--2 hours on a Colab T4, offline), distilled to a fixed grid, and reused across timestep counts and datasets without retraining. After distillation, inference uses only the precomputed grid, just as Uniform, DPM-logSNR, and EDM do. For the transfer studies below we focus on the strongest baselines (DPM-logSNR and EDM); Uniform and AYS are omitted since both are substantially weaker in the matched CIFAR--10 evaluation (Table~\ref{tab:cifar10}).

\subsubsection{Robustness Across Time Grids via Interpolation and Extrapolation}\label{sec:interp}
To test reuse across timestep counts, we take the ART-RL schedule learned on CIFAR--10 at \(K=18\) and construct new grids for \(K\in\{4,9,12,15,20\}\) by log-linear interpolation and extrapolation; EDM and DPM-logSNR use their analytic timestep sequences.

\begin{table}[htbp]
\centering
\caption{CIFAR--10 interpolation/extrapolation FID.}
\label{tab:CIFAR-10-OtherN}
\begin{tabular}{lrrrrr}
\toprule
NFE & 7 & 17 & 23 & 29 & 39 \\
\midrule
DPM-logSNR & 185.63 & 3.52 & 2.19 & 1.94 & 1.88 \\
EDM & 85.80 & 3.11 & 2.06 & 1.88 & 1.85 \\
ART-RL & \textbf{33.73} & \textbf{2.57} & \textbf{2.00} & \textbf{1.84} & \textbf{1.82} \\
\bottomrule
\end{tabular}
\end{table}

Table~\ref{tab:CIFAR-10-OtherN} shows ART-RL preserves the Section~\ref{subsec:cifar} hierarchy across interpolated and extrapolated counts.

\subsubsection{Cross-Dataset Transfer: AFHQv2, FFHQ, and ImageNet}\label{sec:transfer}
We plug the CIFAR--10 ART-RL schedule into the official EDM checkpoints for AFHQv2, FFHQ, and ImageNet, testing a drop-in replacement without dataset-specific retraining.

\begin{figure}[t]
  \centering
  \begin{subfigure}[t]{0.32\linewidth}
    \centering
    \includegraphics[width=\linewidth]{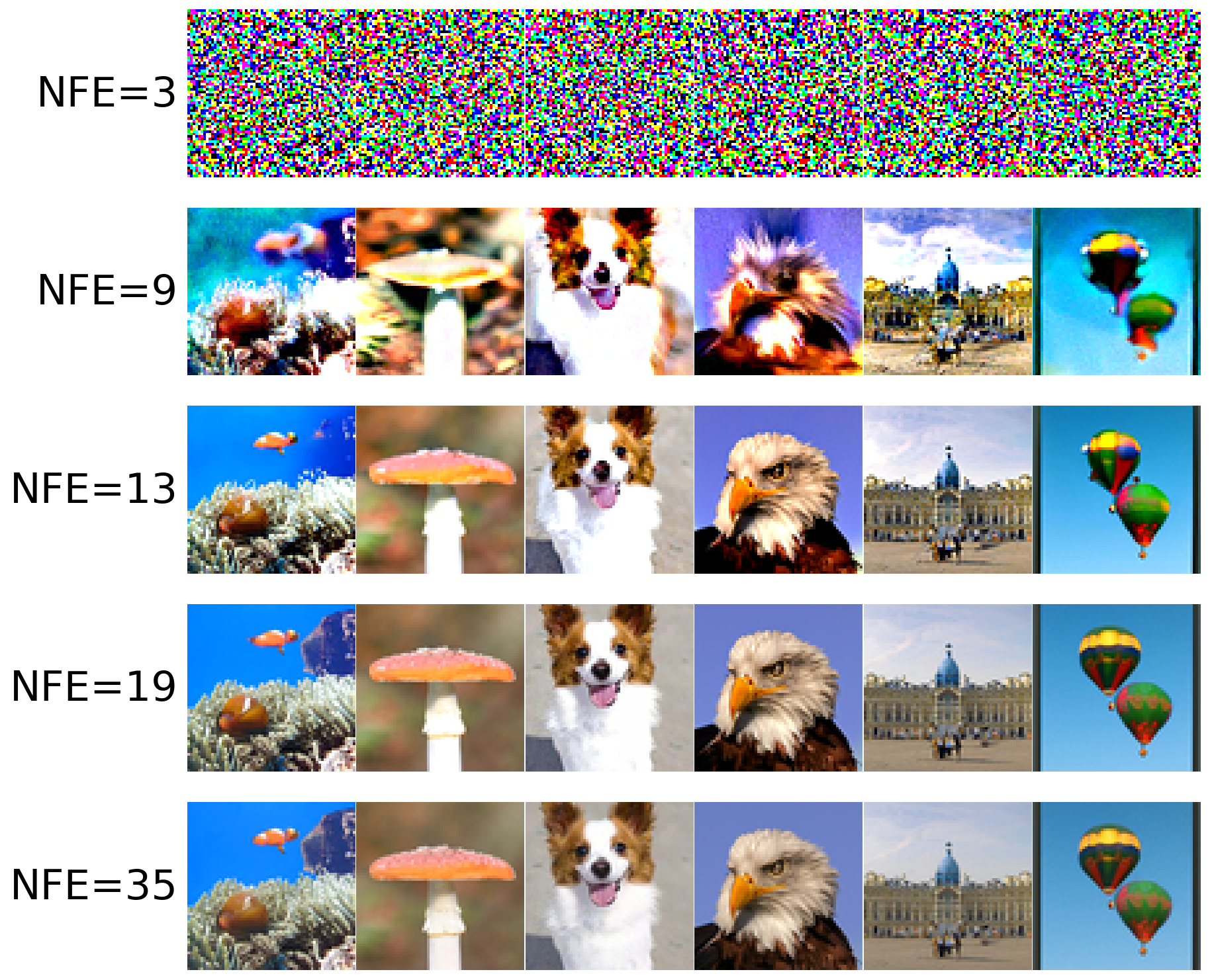}
    \caption{DPM-logSNR}
  \end{subfigure}\hfill
  \begin{subfigure}[t]{0.32\linewidth}
    \centering
    \includegraphics[width=\linewidth]{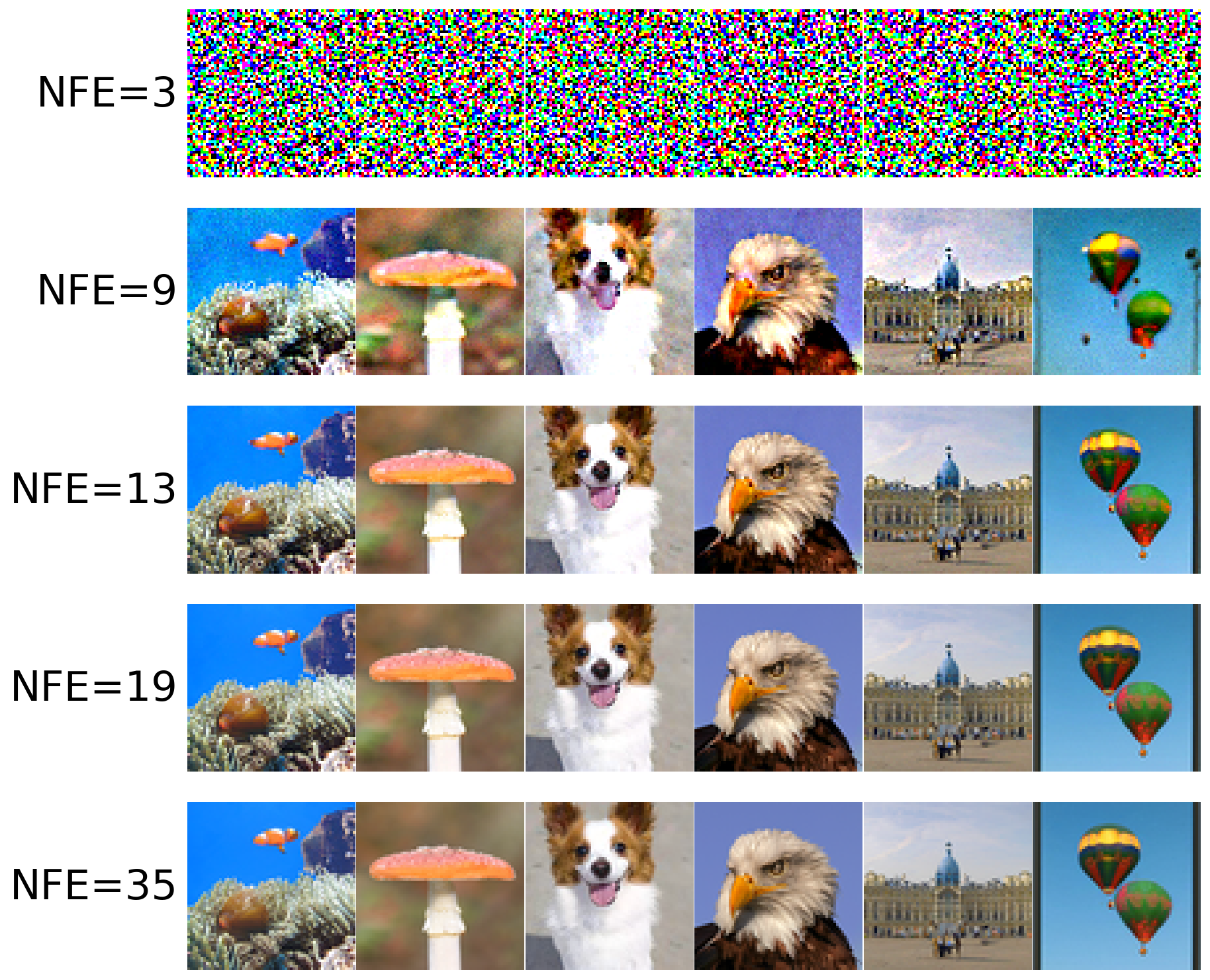}
    \caption{EDM}
  \end{subfigure}\hfill
  \begin{subfigure}[t]{0.32\linewidth}
    \centering
    \includegraphics[width=\linewidth]{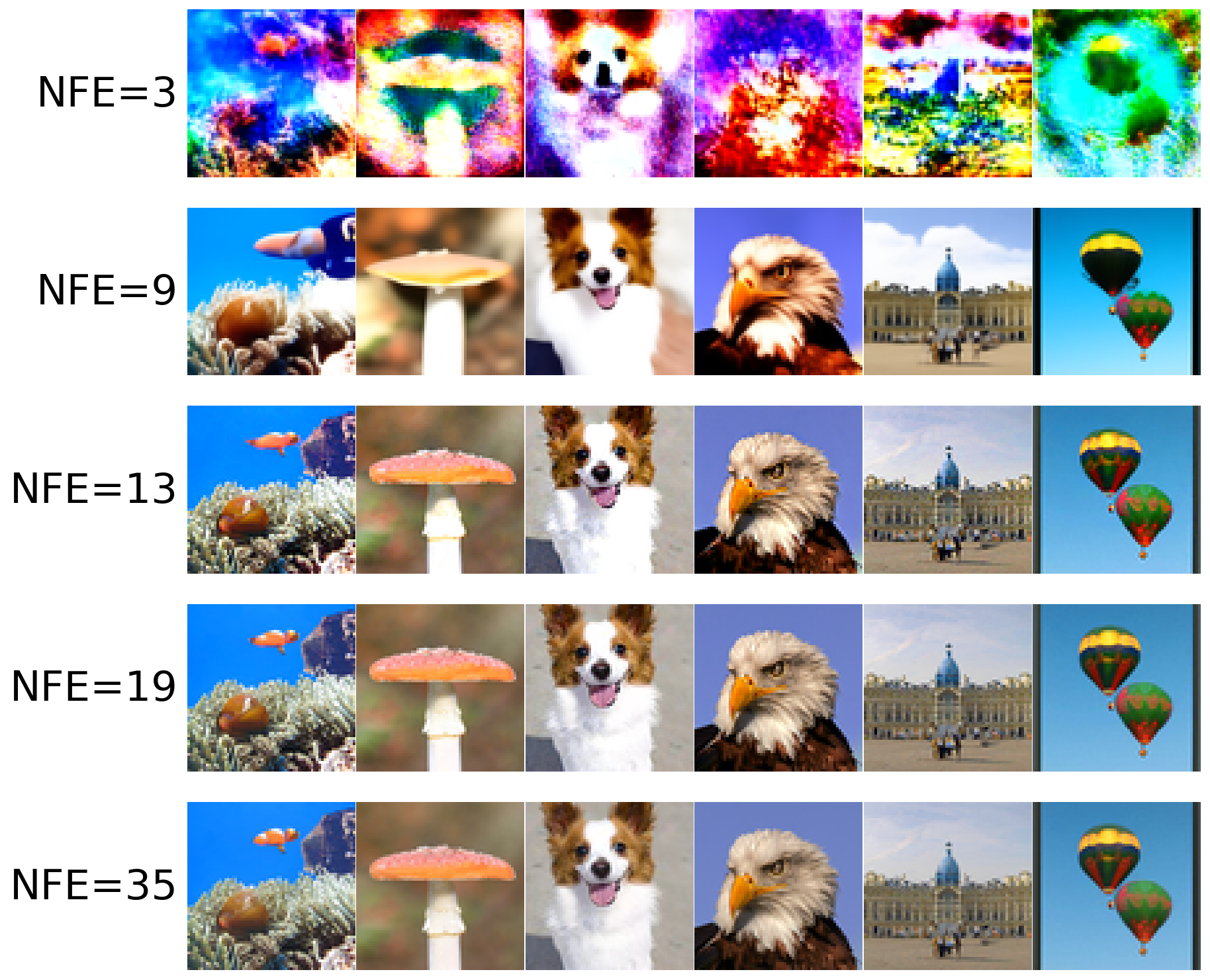}
    \caption{ART-RL}
  \end{subfigure}
  \caption{ImageNet samples across NFEs.}
  \label{fig:imagenet-three-in-a-row}
\end{figure}

\begin{table}[H]
\centering
\caption{Cross-dataset FID using the CIFAR--10 ART-RL schedule.}
\label{tab:cross_dataset_transfer}
\setlength{\tabcolsep}{3.8pt}
\begin{tabular}{llrrrrr}
\toprule
Dataset & Method & 3 & 9 & 13 & 19 & 35 \\
\midrule
\multirow{3}{*}{AFHQv2}
& DPM-logSNR & 375.76 & 67.64 & 9.77 & 3.44 & 2.15 \\
& EDM & 375.76 & 27.88 & 7.56 & 2.99 & 2.11 \\
& ART-RL & \textbf{243.48} & \textbf{20.48} & \textbf{6.12} & \textbf{2.85} & \textbf{2.07} \\
\midrule
\multirow{3}{*}{FFHQ}
& DPM-logSNR & 466.76 & 113.87 & 15.94 & 5.25 & 2.66 \\
& EDM & 466.76 & 57.13 & 15.87 & 5.26 & 2.73 \\
& ART-RL & \textbf{305.97} & \textbf{35.73} & \textbf{11.08} & \textbf{4.31} & \textbf{2.57} \\
\midrule
\multirow{3}{*}{ImageNet}
& DPM-logSNR & 437.42 & 60.48 & 12.31 & 4.46 & 2.66 \\
& EDM & 437.42 & 35.32 & 8.18 & 3.68 & 2.57 \\
& ART-RL & \textbf{147.21} & \textbf{29.49} & \textbf{7.01} & \textbf{3.62} & \textbf{2.56} \\
\bottomrule
\end{tabular}
\end{table}

ART-RL improves over EDM and DPM-logSNR at low and moderate NFEs across all three datasets (Figure~\ref{fig:imagenet-three-in-a-row} and further visual comparisons in \citet{huang2026full}), confirming a reusable rather than dataset-specific parameterization.

\section{Conclusions}
\label{sc6}

We introduced ART for timestep allocation in score-based diffusion models. Its main modeling step is to replace direct grid selection by a continuous time change whose rate is a learnable control that redistributes computation under a fixed time budget. ART-RL solves this control problem through an auxiliary randomized formulation: sufficiency and necessity theorems show that the deterministic ART optimum and the mean of any optimal Gaussian auxiliary policy coincide, making actor--critic learning a principled route to a deterministic schedule. Empirically, within matched EDM samplers, the distilled ART-RL grid improves FID over Uniform, DPM-logSNR, and EDM on CIFAR--10, generalizes across step counts, and transfers without retraining to AFHQv2, FFHQ, and ImageNet as a drop-in time grid with no inference overhead.

Future directions include extending ART to stochastic samplers, designing higher-order or solver-aware error surrogates, and identifying settings where state-dependent schedules improve over distilled time-only grids. More broadly, adaptive time reparameterization for diffusion sampling remains nascent, and ART with ART-RL offers a control-theoretic starting point for systematic schedule design in generative diffusion models.

\acksection
\paragraph{Funding.}
Yilie Huang acknowledges financial support from the Start-up Fund of The Hong Kong Polytechnic University (Project ID: P0063874).
Wenpin Tang is supported by NSF CAREER Award DMS-2538791 and NSF MFAI Award DMS-2602038.
Xun Yu Zhou is supported by the Nie Center for Intelligent Asset Management and the Yao Artificial Intelligence Research Fund at Columbia University.

\paragraph{Competing Interests.}
The authors declare no competing interests.

\bibliographystyle{plainnat}
\bibliography{ref}

\begin{thebibliography}{43}
\providecommand{\natexlab}[1]{#1}
\providecommand{\url}[1]{\texttt{#1}}
\expandafter\ifx\csname urlstyle\endcsname\relax
  \providecommand{\doi}[1]{doi: #1}\else
  \providecommand{\doi}{doi: \begingroup \urlstyle{rm}\Url}\fi

\bibitem[Benton et~al.(2024)Benton, De~Bortoli, Doucet, and
  Deligiannidis]{Ben24}
Joe Benton, Valentin De~Bortoli, Arnaud Doucet, and George Deligiannidis.
\newblock Nearly $ d $-linear convergence bounds for diffusion models via
  stochastic localization.
\newblock In \emph{International Conference on Learning Representations}, 2024.

\bibitem[Chen et~al.(2023{\natexlab{a}})Chen, Lee, and Lu]{Chen2023}
Hongrui Chen, Holden Lee, and Jianfeng Lu.
\newblock Improved analysis of score-based generative modeling: User-friendly
  bounds under minimal smoothness assumptions.
\newblock In \emph{International Conference on Machine Learning}, pages
  4735--4763, 2023{\natexlab{a}}.

\bibitem[Chen et~al.(2023{\natexlab{b}})Chen, Chewi, Li, Li, Salim, and
  Zhang]{ChenChe23}
Sitan Chen, Sinho Chewi, Jerry Li, Yuanzhi Li, Adil Salim, and Anru~R Zhang.
\newblock Sampling is as easy as learning the score: theory for diffusion
  models with minimal data assumptions.
\newblock In \emph{International Conference on Learning Representations},
  2023{\natexlab{b}}.

\bibitem[Gao et~al.(2025)Gao, Zha, and Zhou]{GZZ24}
Xuefeng Gao, Jiale Zha, and Xun~Yu Zhou.
\newblock Reward-directed score-based diffusion models via q-learning.
\newblock \emph{Journal of Machine Learning Research}, 26\penalty0
  (302):\penalty0 1--46, 2025.

\bibitem[{Google}(2024)]{Veo2024}
{Google}.
\newblock State-of-the-art video and image generation with {Veo} 2 and {Imagen}
  3.
\newblock
  \url{https://blog.google/technology/google-labs/video-image-generation-update-december-2024/},
  2024.
\newblock Accessed: 2025-09-17.

\bibitem[Ho et~al.(2020)Ho, Jain, and Abbeel]{Ho20}
Jonathan Ho, Ajay Jain, and Pieter Abbeel.
\newblock Denoising diffusion probabilistic models.
\newblock In \emph{Advances in Neural Information Processing Systems},
  volume~33, pages 6840--6851, 2020.

\bibitem[Huang et~al.(2025{\natexlab{a}})Huang, Huang, and Lin]{HHL25}
Daniel~Zhengyu Huang, Jiaoyang Huang, and Zhengjiang Lin.
\newblock Convergence analysis of probability flow {ODE} for score-based
  generative models.
\newblock \emph{arXiv preprint arXiv:2404.09730}, 2025{\natexlab{a}}.

\bibitem[Huang(2025)]{huang2025alm}
Yilie Huang.
\newblock Continuous-time reinforcement learning for asset--liability
  management.
\newblock In \emph{ACM International Conference on AI in Finance}, pages
  360--368, 2025.

\bibitem[Huang and Zhou(2026{\natexlab{a}})]{huang2025data}
Yilie Huang and Xun~Yu Zhou.
\newblock Data-driven exploration for a class of continuous-time indefinite
  linear--quadratic reinforcement learning problems.
\newblock \emph{IEEE Transactions on Automatic Control}, 2026{\natexlab{a}}.
\newblock \doi{10.1109/TAC.2026.3734857}.
\newblock Early Access.

\bibitem[Huang and Zhou(2026{\natexlab{b}})]{huang2026aid}
Yilie Huang and Xun~Yu Zhou.
\newblock Amortized guidance for image inpainting with pretrained diffusion
  models.
\newblock In \emph{Advances in Neural Information Processing Systems},
  2026{\natexlab{b}}.
\newblock To appear; arXiv:2605.13010.

\bibitem[Huang et~al.(2022)Huang, Jia, and Zhou]{huang2022achieving}
Yilie Huang, Yanwei Jia, and Xun~Yu Zhou.
\newblock Achieving mean--variance efficiency by continuous-time reinforcement
  learning.
\newblock In \emph{ACM International Conference on AI in Finance}, pages
  377--385, 2022.

\bibitem[Huang et~al.(2025{\natexlab{b}})Huang, Jia, and
  Zhou]{huang2025sublinear}
Yilie Huang, Yanwei Jia, and Xun~Yu Zhou.
\newblock Sublinear regret for a class of continuous-time linear-quadratic
  reinforcement learning problems.
\newblock \emph{SIAM Journal on Control and Optimization}, 63\penalty0
  (5):\penalty0 3452--3474, 2025{\natexlab{b}}.

\bibitem[Huang et~al.(2026{\natexlab{a}})Huang, Jia, and Zhou]{huang2024mean}
Yilie Huang, Yanwei Jia, and Xun~Yu Zhou.
\newblock Mean--variance portfolio selection by continuous-time reinforcement
  learning: Algorithms, regret analysis, and empirical study.
\newblock \emph{Management Science}, 2026{\natexlab{a}}.
\newblock To appear; arXiv:2412.16175.

\bibitem[Huang et~al.(2026{\natexlab{b}})Huang, Tang, and Zhou]{huang2026full}
Yilie Huang, Wenpin Tang, and Xun~Yu Zhou.
\newblock Adaptive reparametrized time for score-based diffusion sampling.
\newblock \emph{Journal of Machine Learning Research}, 2026{\natexlab{b}}.
\newblock To appear; arXiv:2607.02137.

\bibitem[Jia and Zhou(2022{\natexlab{a}})]{jia2021policy}
Yanwei Jia and Xun~Yu Zhou.
\newblock Policy evaluation and temporal-difference learning in continuous time
  and space: A martingale approach.
\newblock \emph{Journal of Machine Learning Research}, 23\penalty0
  (154):\penalty0 1--55, 2022{\natexlab{a}}.

\bibitem[Jia and Zhou(2022{\natexlab{b}})]{jia2021policypg}
Yanwei Jia and Xun~Yu Zhou.
\newblock Policy gradient and actor-critic learning in continuous time and
  space: Theory and algorithms.
\newblock \emph{Journal of Machine Learning Research}, 23\penalty0
  (275):\penalty0 1--50, 2022{\natexlab{b}}.

\bibitem[Jia and Zhou(2023)]{JZ22}
Yanwei Jia and Xun~Yu Zhou.
\newblock $q$-learning in continuous time.
\newblock \emph{Journal of Machine Learning Research}, 24\penalty0
  (161):\penalty0 1--61, 2023.

\bibitem[Karras et~al.(2022)Karras, Aittala, Aila, and
  Laine]{karras2022elucidating}
Tero Karras, Miika Aittala, Timo Aila, and Samuli Laine.
\newblock Elucidating the design space of diffusion-based generative models.
\newblock In \emph{Advances in Neural Information Processing Systems},
  volume~35, pages 26565--26577, 2022.

\bibitem[Khanna et~al.(2025)Khanna, Kharbanda, Li, Varma, Wang, Birnbaum, Luo,
  Miraoui, Palrecha, and Ermon]{KK25}
Samar Khanna, Siddhant Kharbanda, Shufan Li, Harshit Varma, Eric Wang, Sawyer
  Birnbaum, Ziyang Luo, Yanis Miraoui, Akash Palrecha, and Stefano Ermon.
\newblock Mercury: Ultra-fast language models based on diffusion.
\newblock \emph{arXiv preprint arXiv:2506.17298}, 2025.

\bibitem[Lee et~al.(2022)Lee, Lu, and Tan]{LLT22}
Holden Lee, Jianfeng Lu, and Yixin Tan.
\newblock Convergence for score-based generative modeling with polynomial
  complexity.
\newblock In \emph{Advances in Neural Information Processing Systems},
  volume~35, pages 22870--22882, 2022.

\bibitem[Li and Yan(2024)]{LYl24}
Gen Li and Yuling Yan.
\newblock Adapting to unknown low-dimensional structures in score-based
  diffusion models.
\newblock In \emph{Advances in Neural Information Processing Systems},
  volume~37, pages 126297--126331, 2024.

\bibitem[Li et~al.(2024)Li, Wei, Chen, and Chi]{LW23}
Gen Li, Yuting Wei, Yuxin Chen, and Yuejie Chi.
\newblock Towards faster non-asymptotic convergence for diffusion-based
  generative models.
\newblock In \emph{International Conference on Learning Representations}, 2024.

\bibitem[Lu et~al.(2022)Lu, Zhou, Bao, Chen, Li, and Zhu]{lu2022dpm}
Cheng Lu, Yuhao Zhou, Fan Bao, Jianfei Chen, Chongxuan Li, and Jun Zhu.
\newblock {DPM-Solver}: A fast {ODE} solver for diffusion probabilistic model
  sampling in around 10 steps.
\newblock In \emph{Advances in Neural Information Processing Systems},
  volume~35, pages 5775--5787, 2022.

\bibitem[Nie et~al.(2025)Nie, Zhu, You, Zhang, Ou, Hu, Zhou, Lin, Wen, and
  Li]{Nie25}
Shen Nie, Fengqi Zhu, Zebin You, Xiaolu Zhang, Jingyang Ou, Jun Hu, Jun Zhou,
  Yankai Lin, Ji-Rong Wen, and Chongxuan Li.
\newblock Large language diffusion models.
\newblock \emph{arXiv preprint arXiv:2502.09992}, 2025.

\bibitem[{OpenAI}(2024)]{Sora2024}
{OpenAI}.
\newblock Sora: Creating video from text.
\newblock \url{https://openai.com/sora}, 2024.
\newblock Accessed: 2025-09-17.

\bibitem[Ramesh et~al.(2022)Ramesh, Dhariwal, Nichol, Chu, and Chen]{Ramesh22}
Aditya Ramesh, Prafulla Dhariwal, Alex Nichol, Casey Chu, and Mark Chen.
\newblock Hierarchical text-conditional image generation with {CLIP} latents.
\newblock \emph{arXiv preprint arXiv:2204.06125}, 2022.

\bibitem[Rombach et~al.(2022)Rombach, Blattmann, Lorenz, Esser, and
  Ommer]{Rombach2022}
Robin Rombach, Andreas Blattmann, Dominik Lorenz, Patrick Esser, and Bj{\"o}rn
  Ommer.
\newblock High-resolution image synthesis with latent diffusion models.
\newblock In \emph{IEEE/CVF Conference on Computer Vision and Pattern
  Recognition}, pages 10684--10695, 2022.

\bibitem[Sabour et~al.(2024)Sabour, Fidler, and Kreis]{sabour2024align}
Amirmojtaba Sabour, Sanja Fidler, and Karsten Kreis.
\newblock Align your steps: Optimizing sampling schedules in diffusion models.
\newblock In \emph{International Conference on Machine Learning}, pages
  42947--42975, 2024.

\bibitem[Singer et~al.(2023)Singer, Polyak, Hayes, Yin, An, Zhang, Hu, Yang,
  Ashual, and Gafni]{Singer2022}
Uriel Singer, Adam Polyak, Thomas Hayes, Xi~Yin, Jie An, Songyang Zhang, Qiyuan
  Hu, Harry Yang, Oron Ashual, and Oran Gafni.
\newblock {Make-A-Video}: Text-to-video generation without text-video data.
\newblock In \emph{International Conference on Learning Representations}, 2023.

\bibitem[Song et~al.(2021{\natexlab{a}})Song, Meng, and Ermon]{DDIM}
Jiaming Song, Chenlin Meng, and Stefano Ermon.
\newblock Denoising diffusion implicit models.
\newblock In \emph{International Conference on Learning Representations},
  2021{\natexlab{a}}.

\bibitem[Song and Ermon(2019)]{Song19}
Yang Song and Stefano Ermon.
\newblock Generative modeling by estimating gradients of the data distribution.
\newblock In \emph{Advances in Neural Information Processing Systems},
  volume~32, pages 11918--11930, 2019.

\bibitem[Song et~al.(2021{\natexlab{b}})Song, Sohl-Dickstein, Kingma, Kumar,
  Ermon, and Poole]{song2020score}
Yang Song, Jascha Sohl-Dickstein, Diederik~P Kingma, Abhishek Kumar, Stefano
  Ermon, and Ben Poole.
\newblock Score-based generative modeling through stochastic differential
  equations.
\newblock In \emph{International Conference on Learning Representations},
  2021{\natexlab{b}}.

\bibitem[Sutton and Barto(1998)]{sutton1998reinforcement}
Richard~S Sutton and Andrew~G Barto.
\newblock \emph{Reinforcement learning: An introduction}.
\newblock MIT Press, Cambridge, MA, 1998.

\bibitem[Tang and Zhao(2025)]{TZSS25}
Wenpin Tang and Hanyang Zhao.
\newblock Score-based diffusion models via stochastic differential equations.
\newblock \emph{Statistics Surveys}, 19:\penalty0 28--64, 2025.

\bibitem[Tang and Zhou(2024)]{TZq24}
Wenpin Tang and Xun~Yu Zhou.
\newblock Regret of exploratory policy improvement and $q$-learning.
\newblock \emph{arXiv preprint arXiv:2411.01302}, 2024.

\bibitem[Wang et~al.(2020)Wang, Zariphopoulou, and Zhou]{wang2020reinforcement}
Haoran Wang, Thaleia Zariphopoulou, and Xun~Yu Zhou.
\newblock Reinforcement learning in continuous time and space: A stochastic
  control approach.
\newblock \emph{Journal of Machine Learning Research}, 21\penalty0
  (198):\penalty0 1--34, 2020.

\bibitem[Wu et~al.(2024)Wu, Chen, and Wei]{WCW24}
Yuchen Wu, Yuxin Chen, and Yuting Wei.
\newblock Stochastic {R}unge-{K}utta methods: Provable acceleration of
  diffusion models.
\newblock \emph{arXiv preprint arXiv:2410.04760}, 2024.

\bibitem[Zhang and Chen(2023)]{ZC23}
Qinsheng Zhang and Yongxin Chen.
\newblock Fast sampling of diffusion models with exponential integrator.
\newblock In \emph{International Conference on Learning Representations}, 2023.

\bibitem[Zhang et~al.(2023)Zhang, Song, and Chen]{ZSC23}
Qinsheng Zhang, Jiaming Song, and Yongxin Chen.
\newblock Improved order analysis and design of exponential integrator for
  diffusion models sampling.
\newblock \emph{arXiv preprint arXiv:2308.02157}, 2023.

\bibitem[Zhao et~al.(2023)Zhao, Tang, and Yao]{ZTY23}
Hanyang Zhao, Wenpin Tang, and David~D Yao.
\newblock Policy optimization for continuous reinforcement learning.
\newblock In \emph{Advances in Neural Information Processing Systems},
  volume~36, 2023.

\bibitem[Zhao et~al.(2024)Zhao, Chen, Zhang, Yao, and Tang]{ZC24}
Hanyang Zhao, Haoxian Chen, Ji~Zhang, David~D Yao, and Wenpin Tang.
\newblock Scores as {A}ctions: a framework of fine-tuning diffusion models by
  continuous-time reinforcement learning.
\newblock \emph{arXiv preprint arXiv:2409.08400}, 2024.

\bibitem[Zhao et~al.(2025)Zhao, Chen, Zhang, Yao, and Tang]{ZC25}
Hanyang Zhao, Haoxian Chen, Ji~Zhang, David Yao, and Wenpin Tang.
\newblock Score as {A}ction: Fine tuning diffusion generative models by
  continuous-time reinforcement learning.
\newblock In \emph{International Conference on Machine Learning}, 2025.

\bibitem[Ziebart et~al.(2008)Ziebart, Maas, Bagnell, and
  Dey]{ziebart2008maximum}
Brian~D Ziebart, Andrew~L Maas, J~Andrew Bagnell, and Anind~K Dey.
\newblock Maximum entropy inverse reinforcement learning.
\newblock In \emph{AAAI Conference on Artificial Intelligence}, volume~8, pages
  1433--1438, Chicago, IL, USA, 2008.

\end{thebibliography}

\clearpage

\end{document}